\documentclass[journal]{IEEEtranTIE}
\usepackage{etex}

\usepackage{graphicx}
\usepackage{subfig}
\usepackage{amsmath}
\usepackage{amsthm} 
\usepackage{amssymb}
\usepackage[font=footnotesize]{caption} 
\usepackage{subfig} 
\usepackage[noadjust]{cite} 
\usepackage{color}
\usepackage{algorithm} 
\usepackage{algpseudocode} 
\usepackage{enumerate}
\usepackage[hidelinks,colorlinks=false]{hyperref}
\usepackage[left=0.75in, right=0.75in, top=0.75in, bottom=0.75in]{geometry}
\usepackage{authblk} 
\usepackage{arydshln} 

\usepackage{bm}
\usepackage{tikz}
\usetikzlibrary{calc} 
\usetikzlibrary{shapes} 
\usetikzlibrary{chains}
\usetikzlibrary{fit}
\usetikzlibrary{arrows}
\usetikzlibrary{decorations.text} 
\usetikzlibrary{decorations.markings}
\usetikzlibrary{decorations.pathmorphing} 
\usetikzlibrary{shadows}
\usetikzlibrary{patterns}
\usetikzlibrary{matrix}
\usepackage{pgfplots}
\usepackage[europeanresistors]{circuitikz}
\usepackage[outline]{contour} 
\contourlength{1.5pt}

\newcommand{\vel}{\mathbf{v}}
\newcommand{\p}{\mathbf{p}}
\newcommand{\s}{\mathbf{s}}
\graphicspath{{figures/}}

\usepackage{multirow}
\usepackage{booktabs}

\begin{document}

\title{Sim-to-Real Transfer in Reinforcement Learning for Maneuver Control of a Variable-Pitch MAV}
\author{Zhikun Wang and Shiyu Zhao
\thanks{Z. Wang, and S. Zhao are with the WINDY Lab, Department of Engineering, Westlake University, Hangzhou, China 
{\tt\small {wangzhikun, zhaoshiyu} @ westlake.edu.cn}
This research work was supported by National Natural Science Foundation of China (Grant No.62473320). Corresponding author: Shiyu Zhao}
}
\maketitle

\begin{abstract}
Reinforcement learning (RL) algorithms can enable high-maneuverability in unmanned aerial vehicles (MAVs), but transferring them from simulation to real-world use is challenging. 
Variable-pitch propeller (VPP) MAVs offer greater agility, yet their complex dynamics complicate the sim-to-real transfer. 
This paper introduces a novel RL framework to overcome these challenges, enabling VPP MAVs to perform advanced aerial maneuvers in real-world settings. 
Our approach includes real-to-sim transfer techniques—such as system identification, domain randomization, and curriculum learning to create robust training simulations and a sim-to-real transfer strategy combining a cascade control system with a fast-response low-level controller for reliable deployment. 
Results demonstrate the effectiveness of this framework in achieving zero-shot deployment, enabling MAVs to perform complex maneuvers such as flips and wall-backtracking.

\end{abstract}

\section{Introduction}
In recent years, the field of robotics has witnessed remarkable progress in developing intelligent and agile autonomous systems \cite{antonio2021learning, xie2021adaptive, sun2022aggressive}. 
Unmanned aerial vehicles (MAVs) have particularly emerged as a focal point of research and innovation due to their widespread applications across domains such as surveillance \cite{delmerico2019current}, reconnaissance \cite{petrovski2021application}, disaster management \cite{daud2022applications}, and environmental monitoring \cite{xin2022swarm}. 
Despite these promising applications, achieving advanced aerial maneuvers with MAVs presents substantial challenges for future development.
The complexity of performing sophisticated maneuvers, such as rapid rotational movements and precise positional adjustments, demands a delicate balance between stability, agility, and precision. 
MAVs must be able to adapt to dynamic environments, which requires sophisticated control systems capable of managing these diverse and often conflicting requirements.

Traditional fixed-pitch MAVs have demonstrated effective flight control for various tasks \cite{song2023reaching, antonio2019deep, yang2018the}. 
However, when it comes to performing complex aerobatic maneuvers that require rapid and precise adjustments for large-scale MAVs, fixed-pitch rotors reveal significant limitations.
The fixed pitch MAV relies on the change of rotor speed to vary thrust, which is limited by the rotor's moment of inertia. 
This constraint further restricts its ability to perform rapid changes in both positive and negative thrust, which are essential for advanced maneuvers.
Consequently, the lack of variable pitch control in fixed-pitch systems impedes their ability to adapt to the dynamic requirements of complex aerobatic tasks.

\begin{figure}[!t]
	\centering
	\includegraphics[width= 0.99\columnwidth]{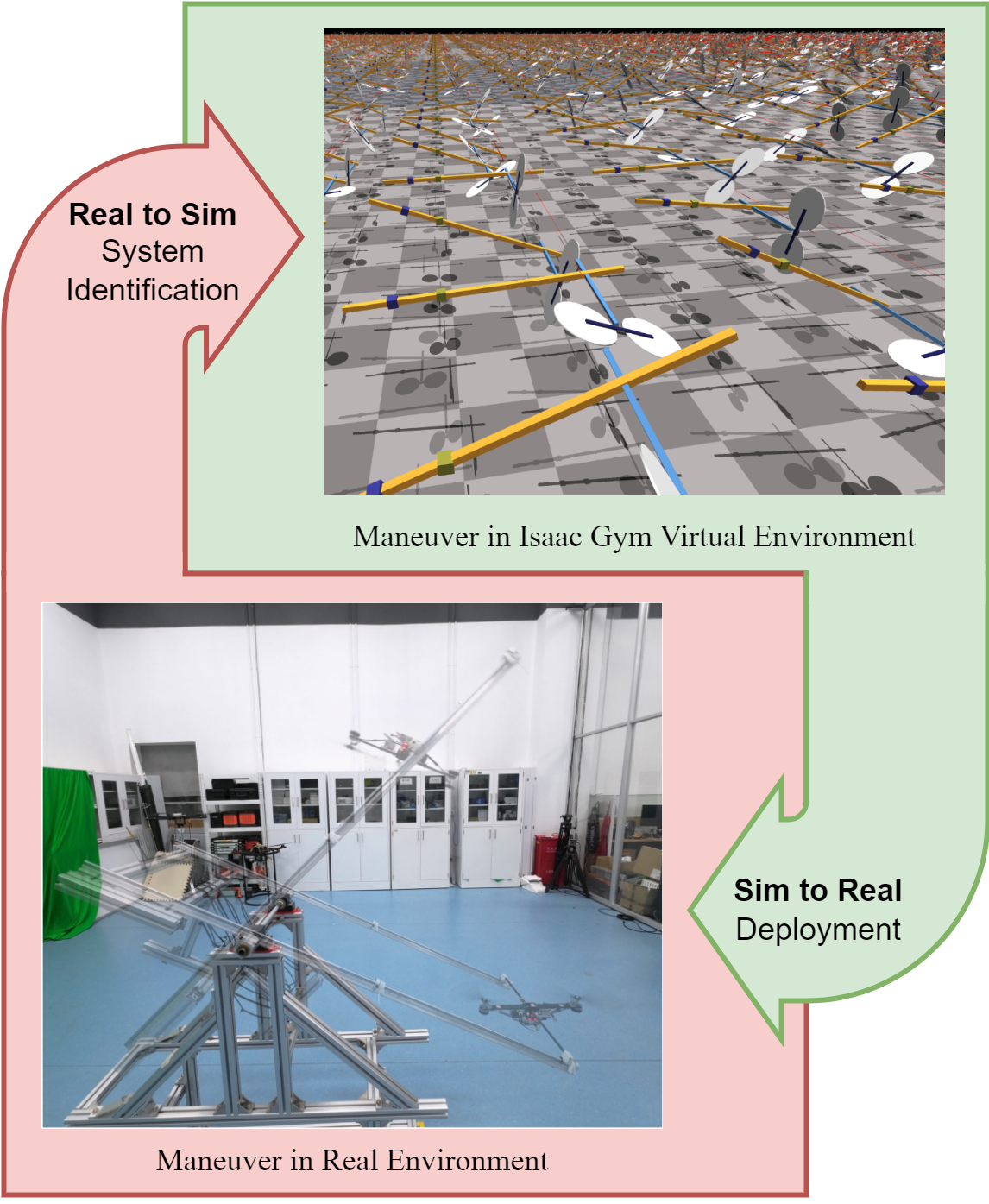}
	\caption{An illustration of the plane variable-pitch MAV achieve maneuver approach. The virtual simulation environment is built with real-world system parameters, and the real-world implementation directly employs the trained controller from the simulation. }
	\label{fig_intro}
\end{figure}

To overcome these limitations, researchers have explored the use of variable-pitch-propeller (VPP) MAVs. 
VPP systems allow for independent adjustment of each actuator’s pitch angle, which significantly enhances the vehicle's maneuverability and control authority \cite{michini2011design, pang2016towards}. 
By enabling dynamic changes in thrust magnitude and magnitude, VPP mechanisms facilitate more complex aerial maneuvers, such as agile flips, precise hovering, and rapid directional changes. 
This increased flexibility allows MAVs to perform advanced aerial feats and adapt to a wider range of operational scenarios. 
However, the integration of VPP systems also introduces additional complexity in the control algorithms required to manage these dynamic capabilities effectively.

Although VPP mechanisms improve MAV control, traditional controllers still face limitations.
The aforementioned studies \cite{cutler2015analysis, cutler2012actuator} have primarily addressed control methods for achieving specific equilibrium states for maneuvers. 
However, current methods is weak facing high maneuvers particularly in handling complex dynamic behaviours and aerodynamic uncertainties. 
Reinforcement learning (RL) has emerged as a powerful tool to overcome these limitations. 
RL trains agents to maximize rewards through trial and error \cite{sutton2018reinforcement} and has proven effective in managing complex, high-dimensional challenges and long planning trajectories \cite{vinyals2019grandmaster, masmitja2023dynamic}. 
The utilization of RL allows us to develop an autonomous learning agent that uses real-time sensory information to adapt and refine its control policies through the process of trials and errors \cite{xu2019learning, hwangbo2019learning, panerati2021learning}.

While RL has proven beneficial in enhancing MAV control, transitioning RL to practical applications with VPP mechanisms presents unique challenges. 
Firstly, it requires the design of a VPP MAV system that can execute rapid and precise pitch angle adjustments, essential for advanced maneuvers. 
Secondly, effective integration demands crafting accurate state representations, reward functions that capture the desired dynamics of complex maneuvers, and controllers capable of coordinating motor adjustments. 
Furthermore, while simulation-based training offers a cost-effective means of data collection, it often involves inherent discrepancies with real-world conditions. 
Our research aims to bridge this gap by transitioning RL-based controllers from virtual simulations to real-world applications, focusing on achieving agile control over VPP MAVs amidst significant aerodynamic interference Fig.~\ref{fig_intro}. 
This includes demonstrating advanced maneuvers, such as flips and wall-backtrack, which showcase significant improvements in maneuver robustness and versatility.

The contributions of this paper are as follows:
\begin{itemize}

\item [1)] Development of an RL Deployment Framework: We developed an RL deployment framework to enable highly agile flight for VPP MAVs. This framework integrates curriculum learning, domain randomization, cascade control systems, and system twins, effectively minimizing the reality gap and ensuring seamless transition from simulation training to real-world performance.  
\item [2)] Creation of a High-Performance VPP MAV Testing Platform: We also created a high-performance testing platform for VPP MAVs, featuring high-frequency hardware, quick-response MAVs, and auxiliary components. This platform significantly enhances thrust control precision and maneuverability, enabling reliable deployment in real-world scenarios.
\end{itemize}

As a result, the proposed controller is successfully deployed to the practical with zero-shot transfer, achieving advanced aerobatic maneuvers such as flips and wall-backtracking. 
This confirms the effectiveness of our proposed framework in enabling agile and precise control for VPP MAVs without the need for extensive prior tuning or adjustment.

\section{Methodology}\label{sec_Problem_Statement}
\subsection{Notation \& System Dynamics}
Due to the fact that the maneuver is confined to a two-dimensional plane, we opted for a planar MAV as the control scheme in order to simplify experimentation \cite{wu2016safety}. 
A variable-pitch propeller actuator is equipped with a servo mechanism capable of adjusting the thrust magnitude of its propellers. 
This innovative design imparts the MAV with enhanced agility and dynamic capabilities.
The planar VPP MAV is modelled as a rectangular cuboid with uniform mass distribution. 

\begin{figure}[!t]
	\centering
	\includegraphics[width= 0.9\columnwidth]{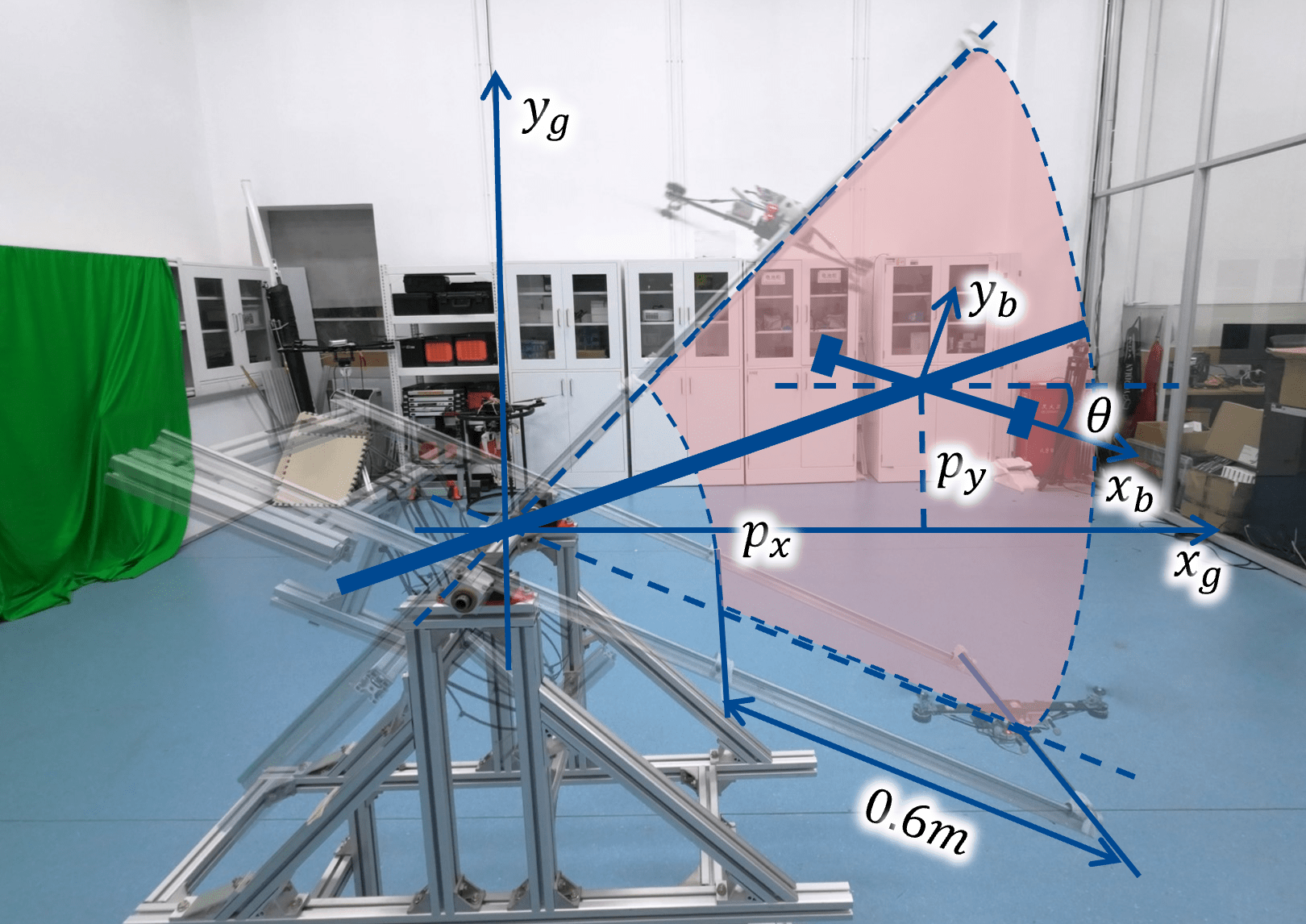}
	\caption{An illustration of the plane variable-pitch MAV and the definition of the coordinate system.}
	\label{fig_illustration}
\end{figure}

The MAV and  the definition of the coordinate system is represented by a 2D dynamic model, as shown in Fig.~\ref{fig_illustration}.
The position and velocity of the centre point of the MAV are $\p \in \mathbb{R}^2$ and $\vel \in \mathbb{R}^2$, respectively. 
The orientation is denoted as $\textbf{R} \in \mathbb{R}^{2 \times 2}$, which is the rotation matrix from the body frame to the global frame, where $\theta \in \mathbb{R}$ is the angle difference between body and world frames.
The angular speed is $q \in \mathbb{R}$.
Let $m$ and $l$ denote the mass and half length of the stick, $I$ the the moment of inertia, $f_i$ is the thrust generated by the $i$-th actuator, $f$ and $\tau$ are the total thrust and torque given by
\begin{equation}\label{forces_merge}
	\begin{split}
		f = & f_1 + f_2,\\
		\tau =& (f_1 - f_2)l.
	\end{split}
\end{equation} 
Then, the overall state vector is $\left[\p,\theta,\vel,q\right] \in \mathbb{R}^6$ and the the dynamic model is
\begin{equation}\label{planar_model}
	\begin{split}
		\dot{\p} &= \vel,\\
		m\dot{\vel} & = -mg \textbf{e}_2 + f\textbf{R} \textbf{e}_2,\\
		\dot{\textbf{R}} &= \textbf{R}[q]_{\times}, \\
		\dot{q} & = -\tau/I,
	\end{split}
\end{equation}
where $g=9.81$ is the gravity acceleration, $\textbf{e}_2 = [0, 1]^T$ and 
\begin{align*}
	\textbf{R} = 
	\begin{bmatrix} 
		\cos \theta & -\sin \theta \\
		\sin \theta & \cos \theta 
	\end{bmatrix}, \quad
	[q]_{\times} = 
	\begin{bmatrix} 
	0 & -q \\
	q & 0
	\end{bmatrix},
\end{align*}
is the planar rotation matrix.
For instance, considering the target position $\p_t$ and angle $\theta_t$, the observed relative position is $\Delta \p = \p_t - \p$ and relative angle is $\Delta \theta = \theta_t - \theta$.

\subsection{Task Formulation}
High-maneuver control of a VPP MAV involves two main objectives: achieving the desired attitude rapidly and preventing the MAV from dropping to the ground. 
During aggressive maneuvers such as flips, the MAV must execute swift orientation changes while maintaining sufficient lift to sustain altitude. 
Intuitively, an agent that completes a maneuver must operate at high speeds and make precise control adjustments, which increases the probability of failure due to the complex dynamics involved. 
Therefore, the optimal maneuver control strategy must define the best trade-off between these competing objectives of agility and stability.

To effectively utilize RL algorithms, we have carefully designed the observation spaces based on the analysis of the system dynamics as
$\s_t = \left[\Delta p_x, \Delta p_y, v_x, v_y, \sin(\Delta\theta), \cos(\Delta\theta), q, \Gamma p_x,\Gamma p_y\right] \in \mathbb{R}^9.$
These observations represent the MAV's relative position, velocity, relative angle (in both sine and cosine forms), angular velocity and accumulated position residuals, respectively.

By representing the angle $\Delta\theta$ using its sine and cosine components, we avoid discontinuities that occur at the $\pm \pi$ boundaries. 
This approach maps the angle to a connected set of rotation $S^1$, providing a continuous representation that simplifies learning and improves the network's ability to generalize \cite{zhou2019continuity}. 

The position residual integral is defined as $\Gamma p = \sum_{n=0}^t \gamma^n \Delta p(n)$, where $\Delta p(n)$ represents the position error at time step $n$ and $\gamma = 0.9$ is a smoothing coefficient that gradually decreases the influence of past errors, preventing numerical instability over long time intervals.
Including $\Gamma p_x$ and $\Gamma p_y$ in the state vector allows the agent to account for accumulated deviations from the desired trajectory, improving control accuracy.

The action vector corresponds to the control inputs for the MAV and is defined as $ a_t = \left[f, q\right]\in \mathbb{R}^2$, corresponds to the desired total thrust and angular velocity. 
Each action dimension is defined within the range \([-1, 1]\).
These normalized actions are mapped to real-world values based on pre-tested parameters, where thrust commands are mapped to a maximum value of $\pm 10$~N, and angular velocity commands are mapped to a maximum of \( \pm 12 \, \text{rad/s}\).
The mapped values are then processed through the low-level angular velocity control and control allocation algorithms to compute the servo outputs for each propeller.

\subsection{Reward Design}

The primary objective of the network is to ensure that the MAV reaches the desired target point at the correct angle.
A well-designed reward function guides the learning process and effectively shapes the behaviours of the agent.
The reward function is composed of five constituent terms, encompassing position error, orientation error, flight stability, and the integral of position error.
The rationale underlying this reward function is twofold: to facilitate the accurate tracking of the desired position by the aerial vehicle and simultaneously enhance the overall stability of its motion. 
\begin{itemize}
	\item [1)] The first component of the reward function is the position reward. To strongly penalize the reward if the position error is too far away from the reference point, the position reward is given by $r_p = 1/(1+ 10||\Delta p||_2^2)$.
	\item [2)] The orientation reward is intended to minimize the angle difference between the current and target orientations. Specifically, since it's a planar maneuver, the angle difference should be less than $\pi$. Otherwise, the system will be encouraged to flip from the other side. The orientation reward is defined as $r_o = (1+ \cos(\Delta\theta))/2$.
	\item [3)] The stability reward maintains steadiness at the target position. It is determined as $r_v = 1/(1+ |v|_2^2)r_p$ and $r_\omega = 1/(1+ ||\omega||_2^2)r_p$ to account for both linear and angular velocity stability.
	\item [4)] The current reinforcement learning input only represents the system's current state, which may result in uncorrected steady-state errors. 
	To detect and correct these long-term errors, we incorporated decaying positional difference integral information into the controller. 
	This integral information accumulates positional differences and gradually decays, helping the controller to identify and correct long-term errors, thereby enhancing the system's stability and accuracy.s The integral of the position residual is given by $r_i = 1/(1+ ||\Gamma p||_2^2)$.
\end{itemize}

In total, the reward design is given as 
\begin{equation}\label{reward_design}
	R = w_pr_p + r_p(w_or_o + w_vr_v + w_\omega r_\omega) + w_ir_i,
\end{equation} 
where $w_p$, $w_p$, $w_v$, $w_\omega$ and $w_i$ are corresponding weights.

\subsection{Policy Training}
We train our agent using the Proximal Policy Optimization (PPO) approach, as reported in \cite{schulman2017proximal}.
A typical PPO Neural Network (NN) comprise two parts, an actor NN and a critic NN.
The actor NN is an agent that works in the environment whereas the critic NN evaluates the performance of the agent.
\subsubsection{Training Environment}

Our reinforcement learning environment is developed based on Isaac Gym, as illustrated in Fig.~\ref{fig_intro}. 
Isaac Gym is a GPU-based parallel robotic simulation environment designed for research and experimentation in robotics and reinforcement learning \cite{makoviychuk2021isaac}. 
It provides parallel physics simulation capabilities, enabling the training and validation of agents across various environments, thereby significantly reducing simulation time.
The overall parallel environments rollouts on up to 8192 which helps to increase the diversity of the collected environment interactions. 

\subsubsection{Asymmetric Network Structure}
We designed different network structures for the actor and critic networks. 
The actor network is optimized for deployment on a microcontroller, which imposes limitations on network size due to its hardware constraints. 
Consequently, the actor network's first layer comprises 96 nodes with the ReLU activation function, while the second layer consists of 64 nodes with the Tanh activation function.
The subsequent layers are fully connected. 
In contrast, the critic network is exclusively used during the training phase and does not require deployment on the microcontroller. 
Therefore, it is designed with a larger network size to leverage more parameters for better performance.
Specifically, the critic network consists of two parts. 
The first part is an LSTM network that compresses information over five time steps. 
The second part is a two-layer MLP, with each layer containing 512 nodes.

\section{Real-to-Sim Transfer}\label{sec_System_Overview}

A major challenge in applying reinforcement learning to real-world scenarios is the real-to-sim gap, which arises from the limitations of simulators in accurately representing reality. 
This gap occurs when actions in the simulation lead to outcomes that differ from what is expected in the real world, resulting in significant discrepancies. 
To bridge this gap, we have employed several strategies.

\begin{figure}[t]
	\centering
	\includegraphics[width= \columnwidth]{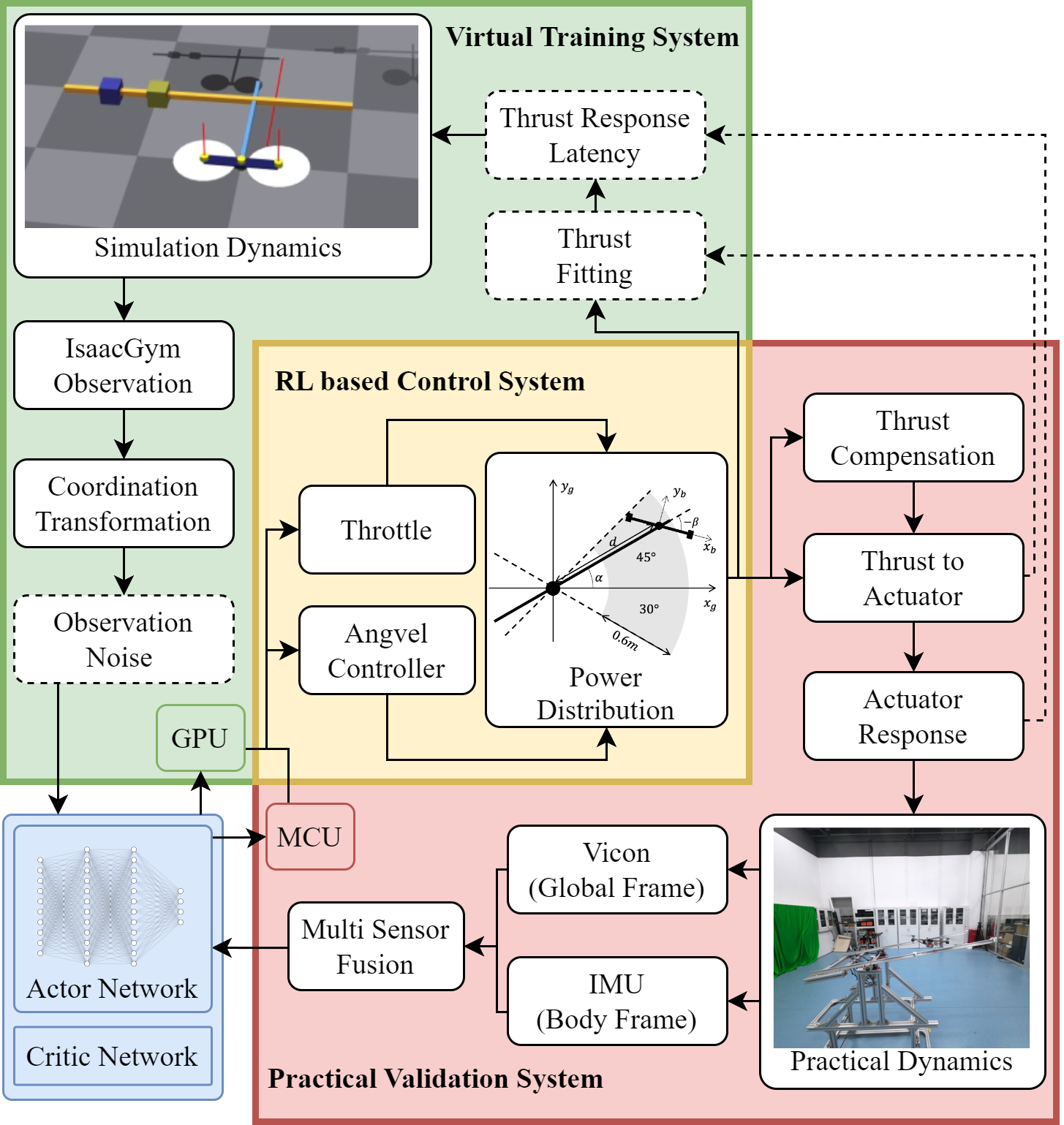}
	\caption{An overview of the proposed VPP MAV control system. }
	\label{fig_full_system}
\end{figure}

\subsection{System Identification}
The most straightforward way is to improve the simulator's realism through careful calibration.
Firstly, we add more detailed simulation to indicate the delays in control signals, and latency in actuator response: 
\begin{itemize}
	\item [1)] Adding a first-order transfer function $G(s) = \frac{1}{Ts+1}$ to servo adjustments, with the parameter determined from the relationship between thrust commands and their response characteristics.
	\item [2)] Incorporating aerodynamic drag $f_d = k_d *|v|*v$ into the dynamics model. The drag coefficient \( k_d \) was estimated using least-squares fitting, based on thrust commands and the corresponding recorded acceleration values.
	\item [3)] Implementing a disturbance model influenced by the propeller angle for motor speed. The variation in motor speed with respect to propeller angle changes was simulated by fitting a polynomial equation to the recorded thrust test data.
\end{itemize}

Secondly, to address the impact of auxiliary equipment on the MAV's movement, this study replicate these facilities within the simulation platform. 
This approach effectively reduces environmental influence.

\subsection{Domain Randomization Methods}
To further enhance the training process, we use domain randomization methods to highly randomize the parameters in simulation in order to cover the real distribution of the real-world data despite the bias between the model and real world:
\begin{itemize}
	\item [1)] Adding noise based on the accuracy of the sensors,
	\item [2)] Randomizing variations in system identification parameters.
\end{itemize}

Observation noise and aerodynamic uncertainty are scaled according to the difficulty level. This scaling allows the network to initially focus on acquiring fundamental control skills in a less noisy and more predictable environment. As the difficulty level increases, the network is gradually exposed to higher levels of noise and uncertainty, promoting the development of robust control strategies capable of handling complex real-world conditions.

Additionally, we employ parameter randomization within the system to broaden the observation and action domains. 
This ensures that the neural network learns the correct relationship between observations and actions, minimizing data-dependent parameter sensitivity.

\subsection{Curriculum Learning}
A dynamic learning rate approach is employed to adjust the learning rate adaptively during training. 
Initially set at 3e-4, the learning rate is maintained for the first 10\% of the total epochs. Subsequently, it is linearly reduced to 9e-5 by the 70\% epoch mark and remains constant at this value until the completion of the training process.

We used curriculum learning method by incorporating a task difficulty level into the training environment to facilitate a progressive learning process. 
Initially set at 0 for the first 10\% of the total training epochs, the difficulty level is linearly increased to 1 by the midpoint of the training and remains constant thereafter. 
During the early stages of training, with difficulty levels below 0.4, tasks are simplified to require only one reward criterion per episode, such as either the position or angle target.

\subsection{Spectral Normalization}
In neural network training and deployment, it is crucial to ensure that the network's output responds to input changes in a stable and controlled manner. 
If the network is overly sensitive to minor input changes, it becomes susceptible to noise and small input errors, which can negatively affect its generalization ability and stability.
To prevent large differences in the output caused by small perturbations in the input, we applied spectral normalization to the actor network. 
Specifically, during training with PPO, we imposed Lipschitz constraints with constant $L$ on all layers. 
By considering the entire network as a composite function $f$ mapping from state 
$\mathbf{s}$ to action $\mathbf{a}$, the scaling relationship complies with the following equation:$||f(\mathbf{s}_1) - f(\mathbf{s}_2)||/||\mathbf{s}_1 - \mathbf{s}_2|| \leq L$, 

where $\mathbf{s}_1$ and $\mathbf{s}_2$ are two different states, $||\cdot||$is L2-norm of the Hilbert space $\mathbb{R}$.
Since activation functions like ReLU and Tanh naturally satisfy the 1-Lipschitz condition, applying Lipschitz constraints to the layers of the network ensures that the overall Lipschitz constant is controlled. 
This approach allows us to build a network that is robust to input perturbations, thereby improving the reliability and stability of the policy.

\section{Sim-to-Real Deployment}
In addition to the real-to-sim transfer challenge, deployment presents another issue. 
Real-world conditions, such as aerodynamics, wind turbulence and sensor errors, introduce unpredictable variables that differ from the virtual environment. 
These discrepancies create performance gaps, as trained models may struggle to generalize when dealing with the complex dynamics and physical limitations of real-world scenarios, especially during maneuvers.
To address this, we designed and implemented a comprehensive cascade control strategies to mitigate the sim-to-real challenge.

\begin{figure*}[!t]
	\centering
	\includegraphics[width= 0.95 \linewidth]{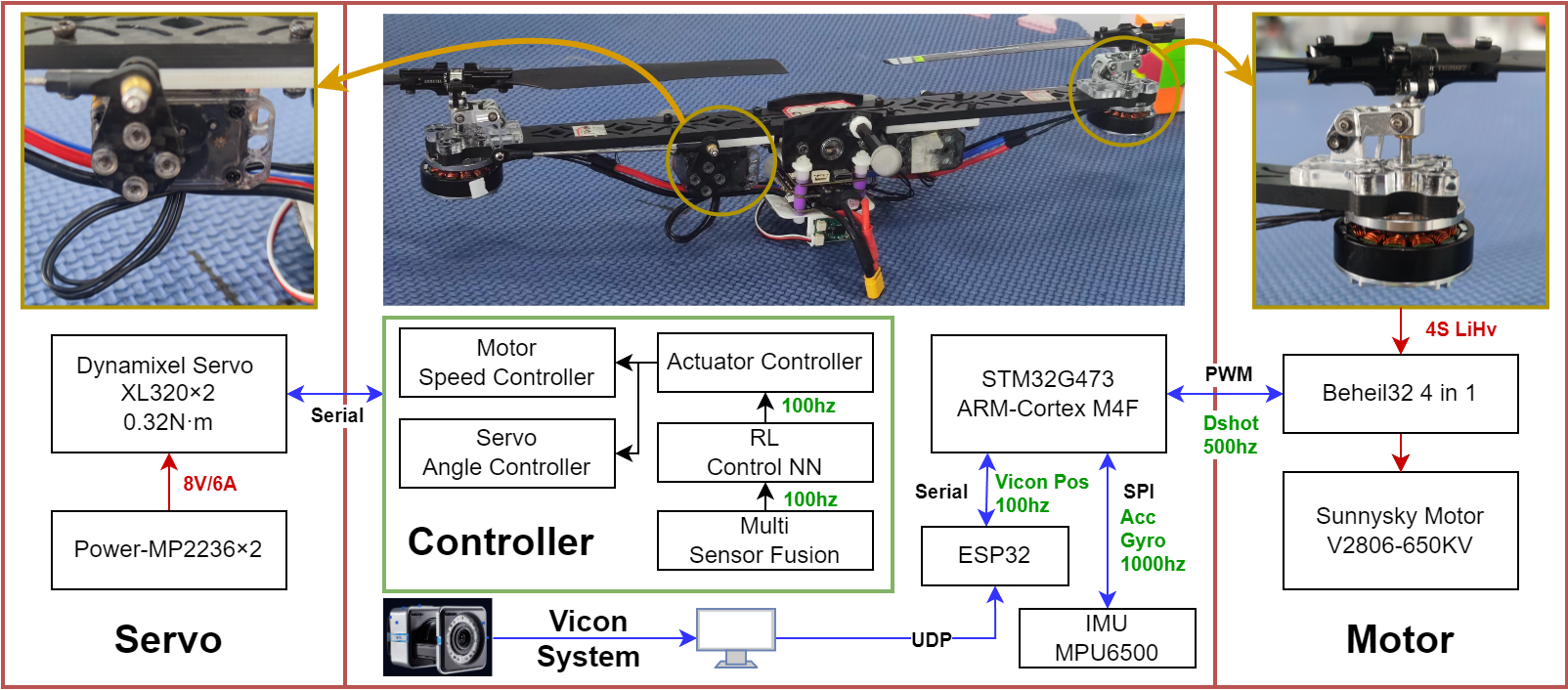}
	\caption{Overview of the components and data flow for the practical planar VPP MAV system. }
	\label{fig_system_diagram}
\end{figure*}

\subsection{System Twin}
To migrate the entire system from simulation to reality, we replicate the same structure in both the simulated and real environments. The system architecture for the VPP MAV control is shown in Fig.~\ref{fig_full_system}.

The system is composed of two main subsystems: the RL training system and the practical validation system. 
Both use the same RL-based hierarchical control system, which includes a neural network controller, a high-frequency angular velocity PD controller, and a power distribution system.

Additionally, apart from the shared system described above, both the virtual training system and the practical validation system have their own unique subsystems.
In the virtual training system, we have developed a sample-based thrust simulation system to fit the thrust response latency in the real-world environment. 
This system aims to mitigate the dynamic response differences between the simulation and reality.
In the validation environment, we have also implemented an additional thrust adaptive control system, and an indoor positioning system utilizing data from multiple sources. 
The indoor positioning system contains a Vicon motion capture system to estimate the position, orientation, and velocity of the VPP MAV, alongside an onboard inertial measurement unit (IMU) for estimating angular velocity.

\subsection{Cascade Control}
The control system is hierarchically organized into three interconnected layers.
At the base is the adaptive actuator control layer, which is responsible for adjusting motor speeds and servo angles to generate the desired thrust in both simulated and real-world environments. 
This layer receives the target forces for each actuator and computes the corresponding motor and servo commands, ensuring accurate actuation and seamless performance across diverse conditions. 

Building on this base, the angular velocity control layer is designed to manage rotational dynamics effectively. 
By processing the target angular velocities, it calculates and outputs the required actuator forces necessary to facilitate precise orientation adjustments. 
This enables the MAV to execute agile flight maneuvers and rapid attitude transitions, which are critical for high-performance operations. 

At the highest level, the system incorporates an RL-based control layer that employs reinforcement learning techniques to optimize control strategies. 
This layer enhances the MAV’s adaptability and decision-making capabilities by processing the current states and desired positions to determine the target angular velocities.
Furthermore, it integrates a perception module for sensor fusion, enabling accurate state estimation by combining data from multiple sensors. 

To ensure robust high-agility control, a Kalman filter was developed to address the challenges of noise and latency inherent in the sensor systems. The IMU provides high-frequency (1000 Hz) measurements of angular velocity and linear acceleration, which are essential for capturing rapid dynamics but are prone to noise and drift. Complementing this, the Vicon system delivers highly accurate position and orientation data at a lower frequency (100 Hz), offering stability and precision but with limited responsiveness to fast dynamics. The Kalman filter combines these complementary data sources, balancing responsiveness and accuracy to provide reliable state estimation. 

To seamlessly transition the entire system from simulation to reality, we implemented the same hierarchical structure in both environments. 
The system architecture for the VPP MAV control is illustrated in Fig.~\ref{fig_system_diagram} and comprises two main subsystems: the RL training system and the practical validation system. 
Both subsystems utilize the same RL-based hierarchical control system, which includes a neural network controller for high-level strategy, a high-frequency angular velocity PD controller for precise orientation control, and a power distribution system.

\subsection{Adaptive Actuator Controller} \label{Adaptive_Actuator_Controller}
Secondly, in conventional MAV systems, fixed-pitch propellers generally create a linear relationship between throttle input and motor speed.
Conversely, the mapping of throttle input to motor control for VPP MAVs differs due to the additional complexity introduced by the variable-pitch mechanism. 
A VPP actuator incorporates two controllable components: a motor and a servo. 
Consequently, implementing a low-level adaptive controller specific to VPP actuators is essential.
This adaptive controller is required to modulate throttle input alongside pitch changes to accommodate variations in propeller blade resistance and maintain consistent motor speed.

Given that the VPP actuator constitutes a multiple-input, multiple-output system, our control loop design aims to decouple its two input-output relationships. 
To achieve this, we initiate a series of tests to elucidate the interdependencies among blade angle, rotational speed, and thrust.
Subsequently, we individually model and fit these relationships, thereby deriving the varying rotational speeds under different operational conditions. 
By utilizing the derived thrusts, we employ forward control to stabilize the actuator's rotational speed.
This is seamlessly integrated with a rotational speed feedback control mechanism, ensuring the actuator's stability and effective management of diverse operational demands.

\begin{figure}[!t]
	\centering
	\subfloat[]{
		\begin{minipage}[t]{0.95\linewidth}
			\centering
			\includegraphics[width=\columnwidth]{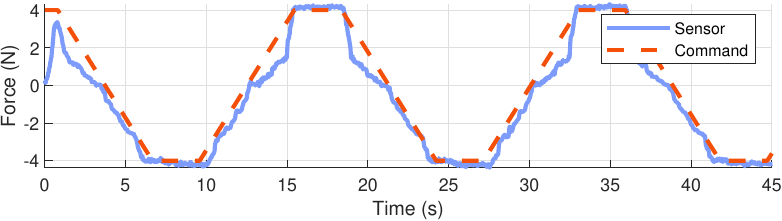}
			\label{fig_adjust_before}
			\vspace{-3em}
		\end{minipage}
	}
	
	\subfloat[]{
		\begin{minipage}[t]{0.95\linewidth}
			\centering
			\includegraphics[width=\columnwidth]{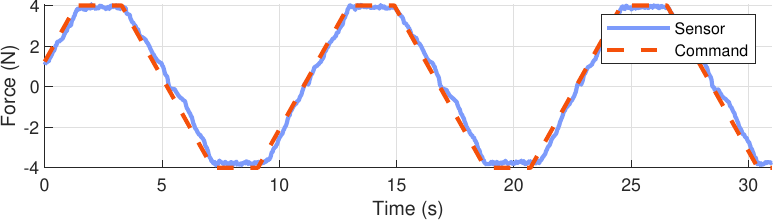}
			\label{fig_adjust_after}
			\vspace{-3em}
		\end{minipage}
	}
	\caption{The experimental results compare the tracking of target thrust by a VPP actuator using different methods, where the commanded thrust is represented by the red dotted line and the response thrust is depicted by the blue solid lines. \protect\subref{fig_adjust_before} shows the thrust response with polynomial regression, whereas \protect\subref{fig_adjust_after} shows the thrust response with adaptive actuator controller. }	
	\label{fig_thrust_compensation}
\end{figure}

Based on actual tests of thrust, speed, and blade angle, we found that the actuator produces significant vibrations and a decrease in thrust when the speed and blade angle are too large.
To ensure that the thrust meets the algorithm requirements, we tested the motor's operating range of the MAV and found that the current weight of the MAV requires the motor to maintain at least 4,000~RPM. 
Also, when the speed exceeds 5,000~RPM, the blade vibration increases rapidly. 
Therefore, we used 4,500~RPM as the target speed for control and tested and fitted the formula between the speed, servo angle, and thrust.

Assuming the propeller speed is almost steady, the thrust $T$ and drag $D$ of propellers are modelled by \cite{cutler2015analysis}
\begin{align}
	T &= k_{T} \omega \alpha, \\
	D &= k_{D1} \omega^2 + k_{D2}\omega^2\alpha^2 + k_{D3}\omega\alpha \label{eq_propeller_drag}, 
\end{align}
where $\alpha$ is the propeller pitch angle, $\omega$ is the propeller rotating speed, $k_{T}, k_{D1}, k_{D2}$ and $k_{D3}$ are constants relating to the physical and aerodynamic properties of the propellers.

The motor-propeller equation can be modelled by
\begin{align}
	I\dot{\omega} &= \left[ \left( V-\omega/k_V \right )\frac{1}{R}-i_0 \right]\frac{1}{k_Q}-D \label{eq_motor_propeller}, \\
	i & =  \left( V-\omega/k_V \right )\frac{1}{R} \label{eq_current},
\end{align}
where $V$ is the voltage applied to the motor, $i$ is the current, $k_V, k_Q, R$ and $i_0$ are the motor related constants.
Substituting Eq.~(\ref{eq_propeller_drag}) and Eq.~(\ref{eq_current}) to Eq.~(\ref{eq_motor_propeller}) gives
\begin{align*}
	I\dot{\omega} = \left(i-i_0 \right)\frac{1}{k_Q} - k_{D1} \omega^2 - k_{D2}\omega^2\alpha^2 - k_{D3}\omega\alpha,
\end{align*}
which reduces to
\begin{align}
	\alpha = \sqrt{\frac{I\dot{\omega}-\left(i-i_0 \right)\frac{1}{k_Q}+ k_{D1}\omega^2 }{-k_{D2}\omega^2}+\left(\frac{k_{D3}}{2k_{D2}\omega}\right)}-\frac{k_{D3}}{2k_{D2}\omega}.\label{eq_alpha}
\end{align}

According to Eq.~(\ref{eq_alpha}), we could find that propeller pitch angle is related to $i, \omega$ and $\dot{\omega}$. 
Recall the assumption that the propeller is working at an almost constant speed which gives $$\dot{\omega} = 0$$ and $\omega$ is a constant, we can simplify Eq.~(\ref{eq_alpha}) to
\begin{align}
	\alpha = \sqrt{g_0(i-i_0) + g_1} - g_2.
\end{align}
Therefore, by measuring the current motor speed and electric current, we can estimate the current thrust using data obtained from pre-tests. 

We first tested the effectiveness of thrust control using a fitting curve designed based on polynomial regression. 
Due to the complexity of the VPP structure and the presence of gaps between structural components during installation, significant discrepancies emerged between thrust and servo control angle after overall assembly, resulting in thrust errors during practical VPP usage, as shown in Fig.~\ref{fig_thrust_compensation}\protect\subref{fig_adjust_before}.
Furthermore, the nonlinear variation in blade resistance due to changes in blade pitch angle also contributes to thrust errors during practical VPP usage.

Subsequently, we evaluated the thrust response of the VPP actuator under the autonomous adaptive thrust algorithm control. 
As depicted in the Fig.~\ref{fig_thrust_compensation}\protect\subref{fig_adjust_after}, the thrust response was notably more accurate.

\section{Experiment}\label{sec_Experiment_Validation}

\subsection{Experiment Setup}
\subsubsection{VPP MAV Hardware}
In this work, the VPP MAV and its components, along with the data flows, are depicted in the middle of Fig.~\ref{fig_system_diagram}. 
The aircraft's frame is crafted from a single 5mm thick carbon fibre board, chosen to achieve both a lightweight structure and reduced vibrations. 
To connect servos and VPP actuators, 3D printing is utilized for producing all necessary connectors.

The pitch control mechanisms, which are designed for small helicopter tail rotors, are depicted on the right side of Fig.~\ref{fig_system_diagram}. 
Servos are positioned away from the actuator, closer to the centre, in order to reduce the total inertia. 
A pushrod is employed to actuate the propeller pitch adjustments.

The deployment process utilized an optimized actor network and FreeRTOS to ensure efficient task management and real-time performance. The actor network, featuring a compact two-layer structure, delivers fast computation and precise control for high-frequency tasks. FreeRTOS facilitated synchronized operation of critical processes, such as RL-based control at 100 Hz and gyro filtering at 1000 Hz. These efficient methods  allows the system to maintain real-time performance without interruptions or delays.

\subsubsection{Plane Support Hardware}
In order to migrate the MAV's maneuver from the three-dimensional plane to two-dimensional space, we attempted to establish an auxiliary device that can ensure that the MAV is limited to movement within a vertical plane while being influenced by the gravitational environment.
This support frame design is constructed by two parallel linear sliding guides and a basic support frame, shown in Fig.~\ref{fig_intro}.
Despite our efforts to balance the mass on both sides of the sliding guides, the remaining inertia caused by its mass cannot be completely eliminated. 
This residual effect may still impact the MAV's trajectory in space, thereby introducing certain discrepancies between its behaviour and true three-dimensional dynamics.
However, it is important to note that the overall mobility of the MAV is not significantly affected.

\subsection{Baseline Comparison on Methods}

\begin{figure}[t]
	\centering
	\includegraphics[width= \columnwidth]{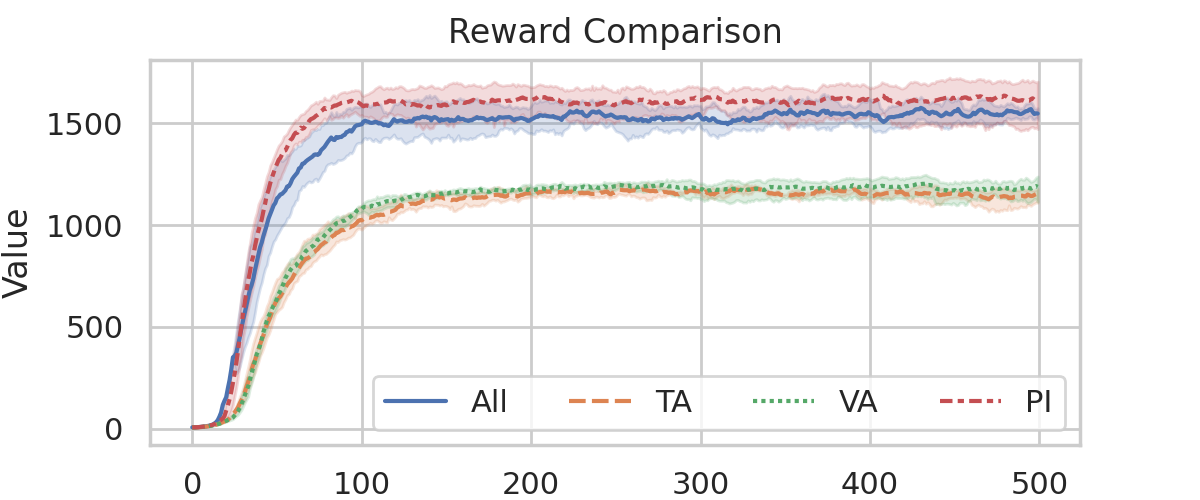}
	\caption{Baseline reward comparison of the proposed methods with basic setup (VA), with trigonometric function angle representation (TA), with position integral information (PI), and with both of them (All). }
	\label{fig_train_compare}
\end{figure}

We first benchmarked the performance of our policy in training environments across four different setups: the basic setup (VA), the setup with trigonometric function angle representation (TA), the setup with position integral information (PI), and the setup incorporating both trigonometric function angle representation and position integral information (All). 
The results are presented in Fig.~\ref{fig_train_compare}.

The data indicates that incorporating position integral information significantly enhances the agent's performance. 
In contrast, the trigonometric function angle representation provides a modest improvement when applied in a visual environment, likely due to the reduced noise and uncertainty in such settings.

Additionally, we evaluated the policy under different initialization conditions, including a broader range of randomizations and noisy situations to better simulate practical scenarios. 
Table~\ref{tab_compare} summarizes the average performance over 100 randomly sampled trajectories. Our method achieved a success rate of 98\% in the target maneuver test, even with noisy and randomized initial states. 
Furthermore, it yielded the lowest position error compared to policies trained with only partial methods or without our proposed methods.
While the randomized initialization approach enhances the drone's ability to handle diverse conditions, it also introduces extreme or challenging scenarios that are inherently difficult to recover from, increasing the likelihood of crashes. Additionally, different RL networks exhibit varying levels of adaptability; some perform well in typical scenarios but struggle under extreme conditions, leading to failures.

\begin{figure*}[!ht]
	\centering
	\includegraphics[width= \textwidth]{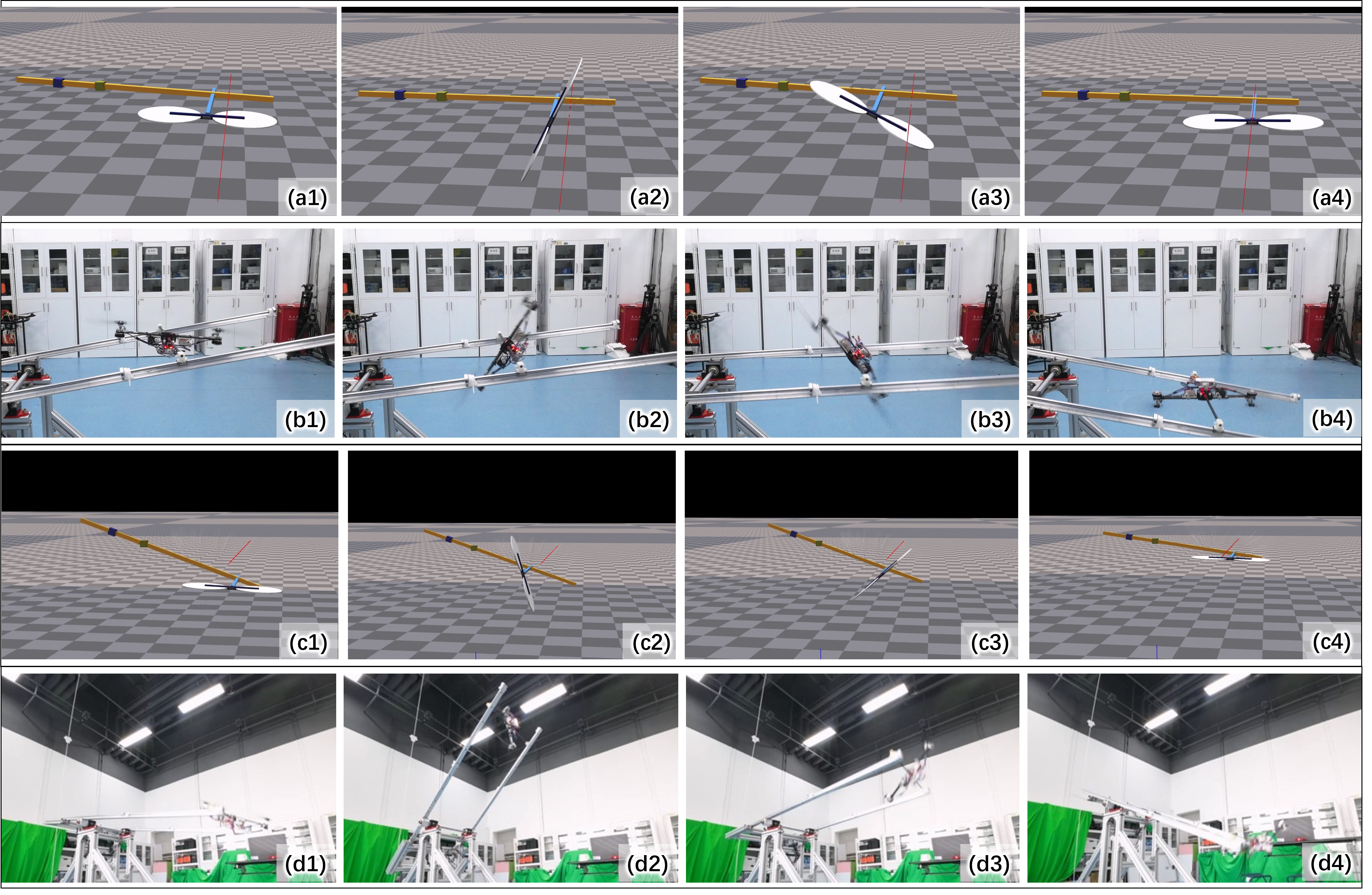}
	\caption{VPP MAV performs maneuvers in the simulation and real-world with the trained NN. 
		(a1) to (a4) illustrate the performance for the planar variable-pitch MAV perform flips from an upward attitude to a downward one in simulated environment, where (b1) to (b4) are the practical images.
		(c1) to (c4) depict the wall-backtrack maneuver in simulated environment, where (d1) to (d4) are the real-world performance.
		The MAV aims to locate at 1.2~m in x axis and 1.25~m in y axis.
	}
	\label{fig_experiment}	
\end{figure*}

\begin{figure*}[!ht]
	\centering
	\includegraphics[width= \textwidth]{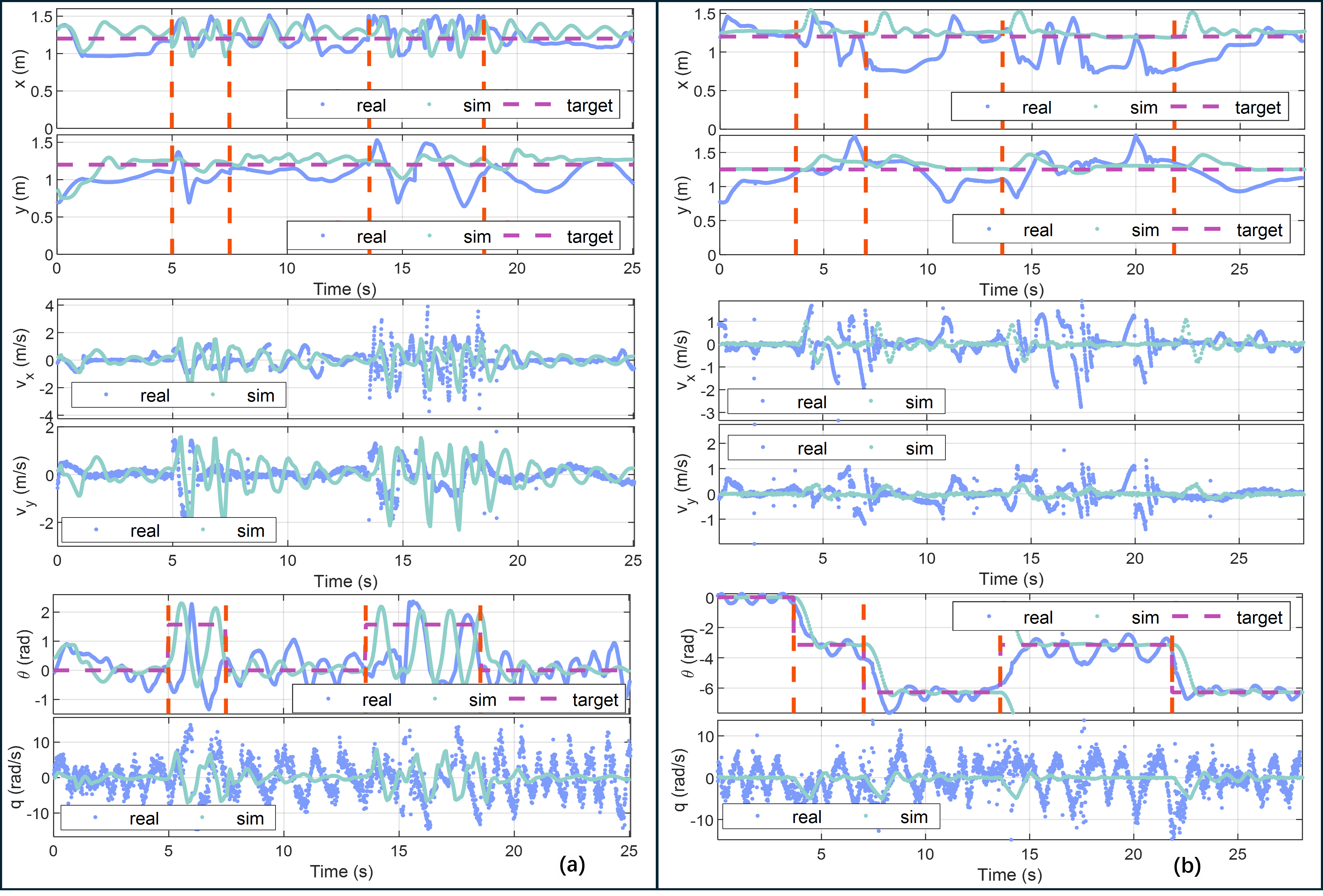}
	\caption{
		Time evolution of the VPP MAV's position, velocity, and rotation in simulation and the real world using the trained neural network. 
		(a) and (b) shows the flight data for the planar variable-pitch MAV perform wall-backtrack and 360-degree flip, respectively.
		The reference trajectory is shown with purple dot lines, while the red dashed lines indicate the moments when the pilot switches the maneuver trigger.
	}
	\label{fig_experiment2}	
\end{figure*}

\begin{table}[t]
	\caption{Performance on randomly generated initial states. We report the tracking error and number of crashes on 100 test tracks.}
	\centering
	\label{tab_compare}
	\begin{tabular}{@{}lrrr@{}}
		\toprule
		Setups & Pos. Err. & Inv. Pos. Err.  & Fail Rate  \\
		\midrule
		VA      & 0.035m   & 0.082m   & 8 \%  \\
		TA 		& 0.038m   & 0.073m   & 8 \%  \\
		PI      & 0.022m   & 0.034m   & 6 \%  \\ 
		All     & 0.017m   & 0.023m   & 2 \% \\
		\bottomrule
	\end{tabular}
\end{table}

\subsection{Real-world Result \& Analysis}
Finally, we test the performance of our policy-generated controller on a practical VPP MAV to verify its transferability.
\subsubsection{Flip Experiment}
In the Flip experiment, the primary goal is to achieve precise positioning and orientation after the flip while minimizing additional movement. Therefore, higher weightings are assigned to position (\(w_p = 0.8\)) and orientation (\(w_o = 0.8\)) to emphasize accuracy. Meanwhile, velocity (\(w_v = 0.2\)) and angular velocity (\(w_\omega = 0.2\)) are assigned lower weightings, reflecting less importance on movement dynamics during the maneuver. Integral error (\(w_i = 0.2\)) is similarly weighted to ensure stability without overemphasising cumulative deviations.

The experimental setup involves commanding the MAV to execute flip maneuvers, as illustrated in the Fig.~\ref{fig_experiment} (b1) to (b4), similar to the simulation scenario shown in Fig.~\ref{fig_experiment} (a1) to (a4).
The complete states, trajectories of the MAV and its corresponding commands are shown in Fig.~\ref{fig_experiment2} (a).

It can be found that there is a steady-state error in the current system, which has two main causes.
Firstly, our practical tests revealed that the lack of torque control, combined with gaps between the slider and the slide rail, causes the slider to get stuck during operation. 
This issue is particularly pronounced when the control outputs are small, as the provided thrust is insufficient to overcome the sticking. 
In these situations, the small control outputs fail to generate enough thrust to free the stuck slider, preventing the complete elimination of the steady-state error.
Secondly, due to the limitations of the microcontroller's performance, we are currently using residual integral as an input to try to reduce this issue. 
Although this method has alleviated the problem to some extent, there is still room for improvement.

\subsubsection{Wall-backtrack Experiment}
The wall-backtrack experiment is designed to evaluate the MAV's performance during a complex aerobatic maneuver. 
This maneuver requires the MAV to transition from a hover position, execute a half-roll to achieve a vertical orientation, and then backtrack to its original hover position.

In the Wall-backtrack experiment, the goal shifts to ensuring the MAV reaches a vertical orientation as a priority while maintaining sufficient position control to avoid ground collisions. Orientation (\(w_o = 1\)) is given the highest weighting to prioritize the vertical alignment. Position (\(w_p = 0.5\)) is moderately weighted to ensure the MAV avoids falling but does not over-constrain its position. Lower weightings for velocity (\(w_v = 0.1\)), angular velocity (\(w_\omega = 0.1\)), and integral error (\(w_i = 0.1\)) reflect a reduced emphasis on these factors, as movement speed and cumulative deviations are less critical in this context. 

Initially, we trained the target platform in a virtual environment but found that the thrust-to-weight ratio was insufficient to support such high-difficulty maneuvers. 
To address this issue, we increased the thrust-to-weight ratio to 2.5 in the simulation, and the results are shown in Fig.~\ref{fig_experiment}(d1) to (d4).
Simultaneously, we directly transferred the same network to the real environment, and the results are depicted in Fig.~\ref{fig_experiment}(e1) to (e4).
The complete states, trajectories of the MAV, and its corresponding commands are presented in Fig.~\ref{fig_experiment2}(b). 
Remarkably, despite the thrust-to-weight ratio limitation in reality, the MAV was able to leverage the elastic properties of its frame to assist in completing the maneuver, albeit with larger positional errors. 
This unexpected behaviour, which was not explicitly trained for in the simulation, highlights the robustness of the RL-based controller.

\section{Conclusion}\label{sec_Conclusion}

This paper presents a purely RL-based controller designed for a planar MAV to execute flip maneuvers. 
We validated the controller's performance by successfully demonstrating flip and wall-backtrack maneuvers in both simulated and real-world environments. 
This research extends the RL algorithm's capabilities from simulation to practical application, providing agile and maneuverable MAVs capable of captivating aerial displays. 
Future work will explore the algorithm's performance in a 3D environment without plane support platform.
\bibliographystyle{Bibliography/IEEEtranTIE}
\bibliography{main}

\end{document}